\documentclass[twocolumn]{fairmeta}
\usepackage{wrapfig}
\usepackage{tabularx}
\usepackage{textcomp}
\usepackage{stfloats}
\usepackage{url}
\usepackage{verbatim}
\usepackage{graphicx}
\usepackage{titlesec}
\usepackage{tocloft}
\usepackage{adjustbox}
\usepackage{enumitem}
\usepackage{multirow}
\usepackage{pifont}
\usepackage{tikz}
\usepackage{comment}
\usepackage{amsmath,amssymb} 
\usepackage{colortbl}  
\usepackage{color}
\usepackage{booktabs} 
\usepackage{hyperref}
\usepackage{graphicx}    
\usepackage{subcaption} 
\usepackage{booktabs}
\usepackage{amsmath} 
\usepackage{multirow} 
\usepackage{booktabs} 
\usepackage{subcaption} 
\usepackage{todonotes}
\RequirePackage{xspace}
\makeatletter
\DeclareRobustCommand\onedot{\futurelet\@let@token\@onedot}
\def\@onedot{\ifx\@let@token.\else.\null\fi\xspace}

\def\ie{\emph{i.e}\onedot}

\makeatother

\definecolor{adptorange}{RGB}{248, 205, 172}
\definecolor{cmpblue}{RGB}{189, 215, 238}
\definecolor{cmpblue}{RGB}{189, 215, 238}

\definecolor{our_red}{RGB}{232,157,160}
\definecolor{our_blue}{RGB}{136,206,230}
\definecolor{our_orange}{RGB}{246,200,168}
\definecolor{our_green}{RGB}{178,211,164}

\definecolor{attn_code0}{RGB}{247,215,200}
\definecolor{attn_code1}{RGB}{238,169,139}
\definecolor{mlp_code0}{RGB}{204,201,221}
\definecolor{mlp_code1}{RGB}{102,95,153}

\definecolor{token_blue}{RGB}{84, 120, 140}

\usepackage{pifont}       
\usepackage{bbding}       
\usepackage{fontawesome}
\usepackage{xspace}

\usepackage{float}

\newlength\savewidth

\newcolumntype{x}[1]{>{\centering\arraybackslash}p{#1pt}}
\newcolumntype{y}[1]{>{\raggedright\arraybackslash}p{#1pt}}
\newcolumntype{z}[1]{>{\raggedleft\arraybackslash}p{#1pt}}

\renewcommand{\paragraph}[1]{\vspace{1mm}\noindent\textbf{#1}}
\usepackage{colortbl}
\usepackage{xcolor}
\usepackage{wrapfig}

\renewcommand{\paragraph}[1]{\vspace{1.25mm}\noindent\textbf{#1}}

\usepackage{algorithm}
\usepackage{listings}

\definecolor{codeblue}{rgb}{0.25, 0.5, 0.5}
\definecolor{codekw}{rgb}{0.35, 0.35, 0.75}
\lstdefinestyle{Pytorch}{
    language = Python,
    backgroundcolor = \color{white},
    basicstyle = \fontsize{9pt}{8pt}\selectfont\ttfamily\bfseries,
    columns = fullflexible,
    aboveskip=1pt,
    belowskip=1pt,
    breaklines = true,
    captionpos = b,
    commentstyle = \color{codeblue},
    keywordstyle = \color{codekw},
}

\definecolor{green}{HTML}{009000}
\definecolor{red}{HTML}{ea4335}

\title{RealisDance-DiT: Simple yet Strong Baseline towards Controllable Character Animation in the Wild}

\author{%
  Jingkai Zhou$^{*1,2,3}$, 
  Yifan Wu$^{*4}$, 
  Shikai Li$^{1,3}$, 
  Min Wei$^{1,3}$, 
  Chao Fan$^{5}$, 
  Weihua Chen$^{\dagger1,3}$,
  Wei Jiang$^{2}$, 
  Fan Wang$^{1,3}$\\
  \small{
  $^1$DAMO Academy, Alibaba Group, 
  $^2$Zhejiang University, 
  $^3$Hupan Lab,
  $^4$Southern University of Science and Technology,
  $^5$Shenzhen University}\\
  \small{$^*$Equal contribution. $^\dagger$ Corresponding author.}\\
  \texttt{zhoujingkai.zjk@alibaba-inc.com, kugang.cwh@alibaba-inc.com}
}
\abstract{
Controllable character animation remains a challenging problem, particularly in handling rare poses, stylized characters, character-object interactions, complex illumination, and dynamic scenes. 
To tackle these issues, prior work has largely focused on injecting pose and appearance guidance via elaborate bypass networks, but often struggles to generalize to open-world scenarios. 
In this paper, we propose a new perspective that, as long as the foundation model is powerful enough, straightforward model modifications with flexible fine-tuning strategies can largely address the above challenges, taking a step towards controllable character animation in the wild. Specifically, we introduce RealisDance-DiT, built upon the Wan-2.1 video foundation model. Our sufficient analysis reveals that the widely adopted Reference Net design is suboptimal for large-scale DiT models. Instead, we demonstrate that minimal modifications to the foundation model architecture yield a surprisingly strong baseline. We further propose the low-noise warmup and ``large batches and small iterations'' strategies to accelerate model convergence during fine-tuning while maximally preserving the priors of the foundation model. In addition, we introduce a new test dataset that captures diverse real-world challenges, complementing existing benchmarks such as TikTok dataset and UBC fashion video dataset, to comprehensively evaluate the proposed method. Extensive experiments show that RealisDance-DiT outperforms existing methods by a large margin. The project page is at this link\footnote{\scriptsize\url{https://thefoxofsky.github.io/project_pages/RealisDance-DiT/index}}.}

\date{\today}

\begin{document}
\maketitle

\begin{figure*}
  \includegraphics[width=\textwidth]{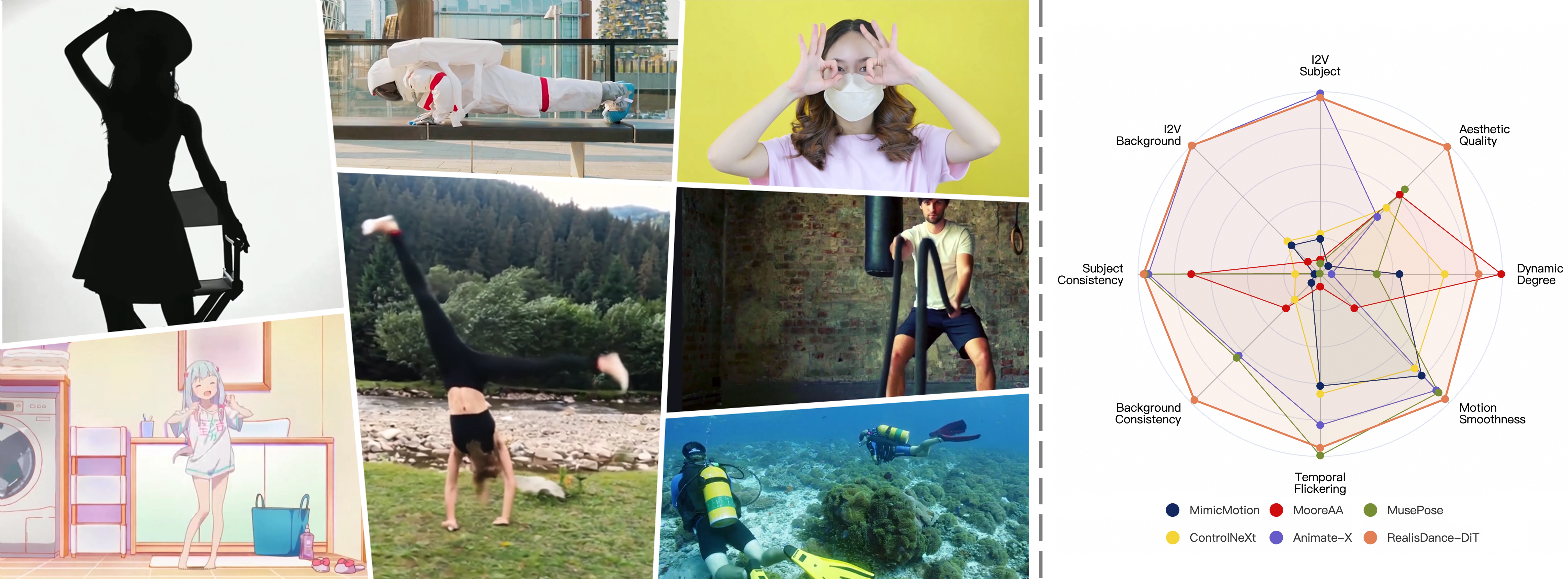}
  \caption{Results of RealisDance-DiT. Left: Frames generated by RealisDance-DiT. Right: Evaluation on the proposed RealisDance-Val dataset using VBench-I2V metrics. Zoom in for better visibility. Please refer to the project page for more videos.}
  \label{fig:teaser}
\end{figure*}

\section{Introduction}

Controllable character animation can be widely applied in film production, virtual digital humans, and e-commerce promotions. This task has recently garnered significant attention from both academia~\cite{disco, magicanimate, animateanyone, animateanyone2, champ, ControlNeXt, animatex, humanvid} and industry~\cite{musepose, mooreaone, realisdance, viggle_ai}, due to advances in generative models and increasing demand for personalized content creation.

\begin{figure}[t]
\includegraphics[width=0.95\linewidth]{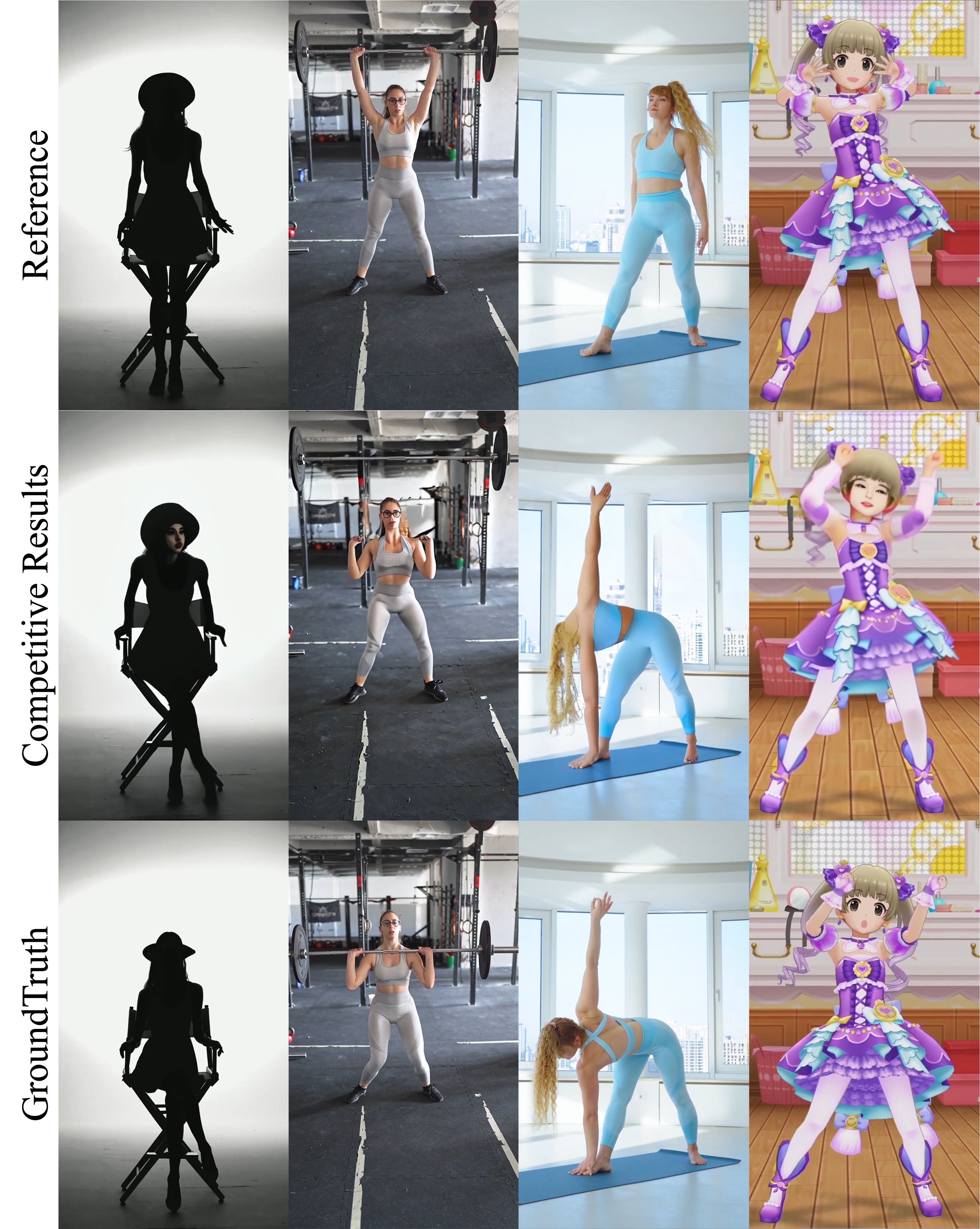}
\centering
\caption{Failure cases of existing methods. Existing methods sometimes generate a face in the silhouette frame, leave dumbbells suspended in the air when the woman squats down, generate artifacts when producing the yoga pose, and generate a realistic face for the comic character.}
\label{fig:issues}
\end{figure}

Existing methods~\cite{animateanyone, magicanimate, animateanyone2, champ, ControlNeXt, animatex, humanvid} employ the Reference Net to inject the reference character ID, achieving significant advances in character consistency. 
However, their performance in open scenes remains unsatisfactory, as they struggle to address challenges such as rare poses, stylized characters, interactions between characters and objects, complex lighting conditions, and scene changes. See Figure~\ref{fig:issues} for example. Existing methods struggle with complex lighting conditions and generate a face artifact in the silhouette frame. Also, it is difficult for existing methods to produce character-object interactions, the generated results leave dumbbells suspended in the air when the woman squats down. In cases of rare poses, existing methods tend to introduce artifacts in body parts where the model lacks adequate understanding. Moreover, when facing stylized characters, existing methods tend to generate incorrect body parts, such as producing a realistic face for the comic character. Intuitively, all of these issues can be attributed to three possible reasons: 1) the Reference Net is not well designed and not robust enough, 2) the main model is not powerful enough to handle these challenges, and 3) the overall fine-tuning is insufficient in terms of both data and iterations.

In this paper, we propose a new perspective that, 
as long as the foundation model is powerful enough, simple model modifications with flexible fine-tuning strategies can unlock the potential for controllable character animation in the wild. Specifically, we propose RealisDance-DiT, which is built upon the strong video foundation model Wan-2.1~\cite{wanxiang}. We only make a few simple adjustments to Wan-2.1, including adding condition input layers and modifying the RoPE position encoding, which yields a superior baseline. 
Our analysis reveals that applying Reference Net does not help with such large DiT foundation models, and even brings negative effects beyond heavy additional overhead. This is because the existing large video foundation model itself possesses the capability to achieve controllable character animation in the wild.  The key lies in guiding the model to unlock this capability, rather than adding additional complex structures.

Given that we utilize a powerful video foundation model, the fine-tuning strategies should prioritize preserving prior knowledge within the model. Therefore, we propose two flexible and effective fine-tuning strategies to accelerate the convergence process while maximally preserving the priors in the foundation model. The first one is the \textit{low-noise warmup} strategy. Our experiments demonstrate that reducing the amount of noise added during the early stages of fine-tuning can speed up convergence. Samples with low added noise are easier for the model to process than those with high added noise.
Using simpler samples for warm-up fine-tuning helps stabilize the adaptation to the new task, rather than starting with difficult samples that may push the model away from the initial local optimum established during pre-training.  The second strategy is called \textit{large batches and small iterations}. We suggest distributing the data in larger batches and fewer iterations. Larger batch sizes enable the model to benefit from more informative yet smooth gradients per update, allowing it to focus on important factors in the downstream task, rather than being hindered by noise in the data. Fewer iterations help the model to keep the pre-trained priors, reducing the risk of overfitting on downstream datasets. Together, these two strategies facilitate faster convergence while preserving rich priors, which is crucial for fine-tuning powerful video foundation models.

In addition, we curated a test dataset named RealisDance-Val, comprising 100 videos with corresponding conditions. This dataset features diverse and challenging scenarios, including rare poses, stylized characters, dynamic scenes, complex lighting conditions, and interactions between characters and objects.  It is specifically designed to evaluate the performance of generative models in open scenes. The proposed method is evaluated on the TikTok dataset~\cite{tiktokdataset}, UBC fashion video dataset~\cite{ubcdataset}, and RealisDance-Val datasets. Experimental results demonstrate that RealisDance-DiT performs favorably against existing methods. Figure~\ref{fig:teaser} exhibits several generated results of RealisDance-DiT. 

In short, we make the following contributions:
\begin{itemize}[itemsep=0mm, leftmargin=*]
    \item Challenged the traditional view that requires heavy reference networks to inject character ID. Instead, we emphasize the powerful video foundation model that naturally processes the capability of character consistency. We only need to bring it out via straightforward modifications.
    \item Proposed two flexible fine-tuning strategies to accelerate convergence while maximally preserving the built-in priors of the powerful video foundation model.
    \item Collected an open-scene test dataset and provided the field with a structurally simple, empirically robust, and experimentally strong baseline model, which is expected to inspire future work.
\end{itemize}

\begin{figure*}[t]
\centering
\tabcolsep=0mm
\begin{tabular}{cccc} 
    \begin{minipage}[b]{0.18\linewidth} 
        \centering
        \includegraphics[height=65mm]{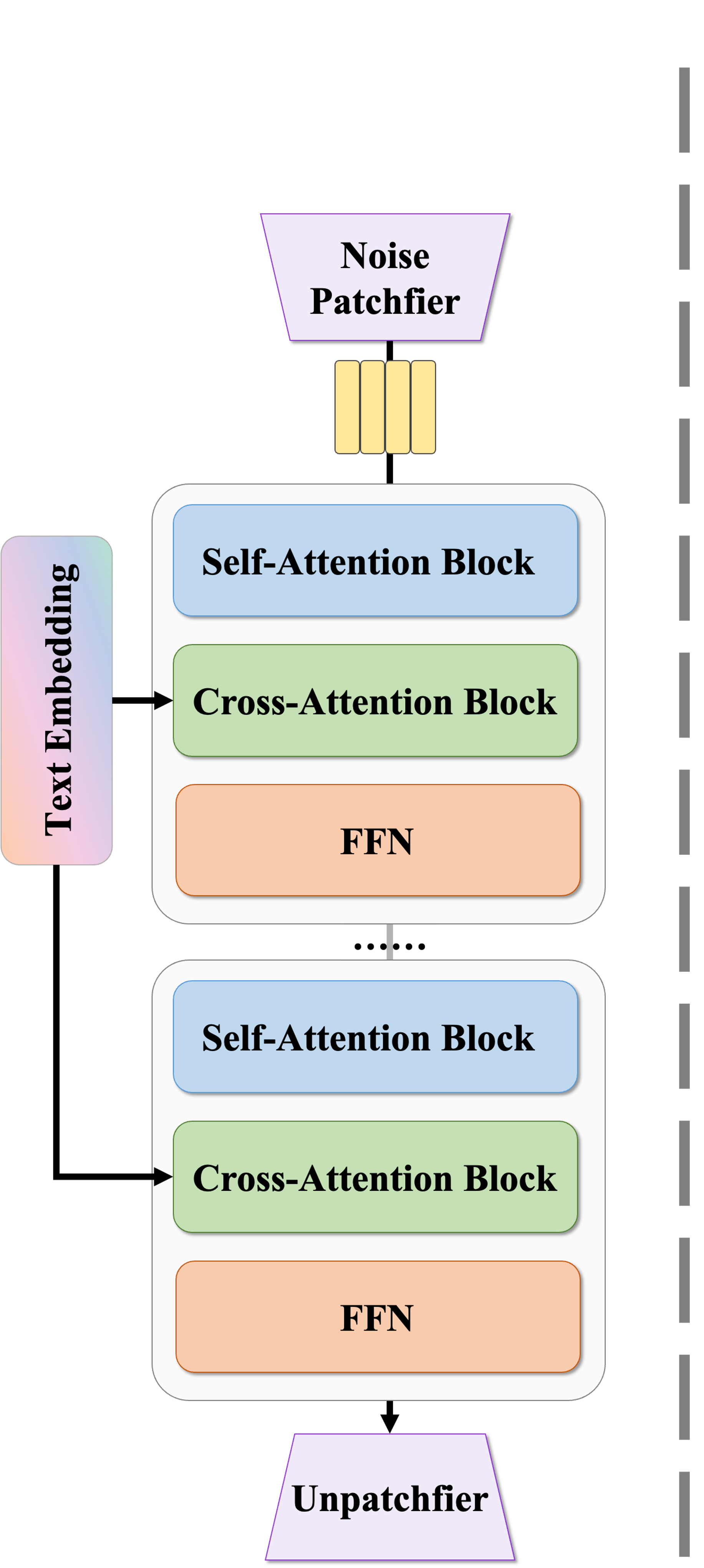} 
        \subcaption{\scriptsize{Original Wan-2.1}}
    \end{minipage} &
    \begin{minipage}[b]{0.36\linewidth} 
        \centering
        \includegraphics[height=65mm]{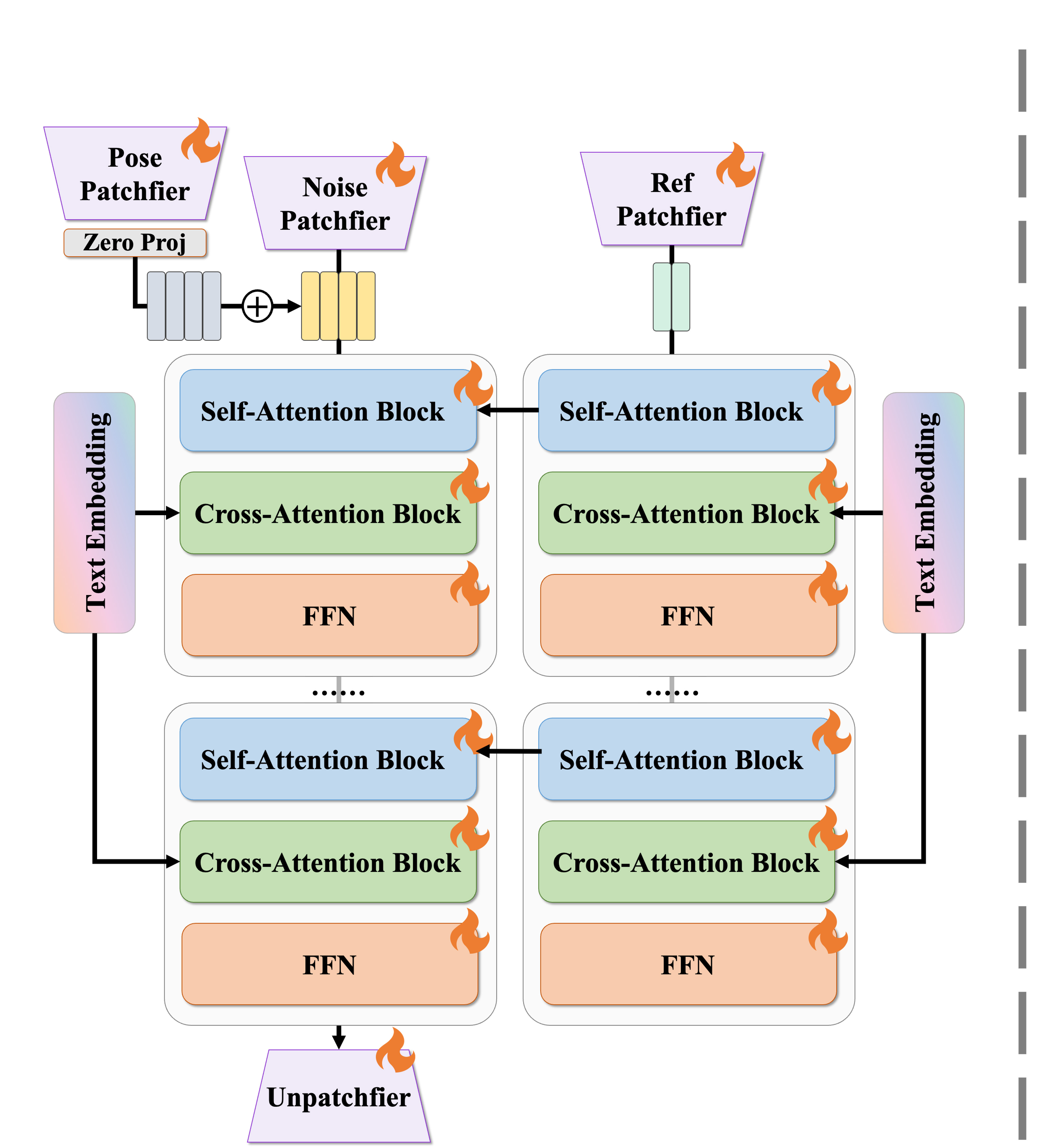} 
        \subcaption{\scriptsize{Reference Net variant}}
    \end{minipage} &
    \begin{minipage}[b]{0.25\linewidth} 
        \centering
        \includegraphics[height=65mm]{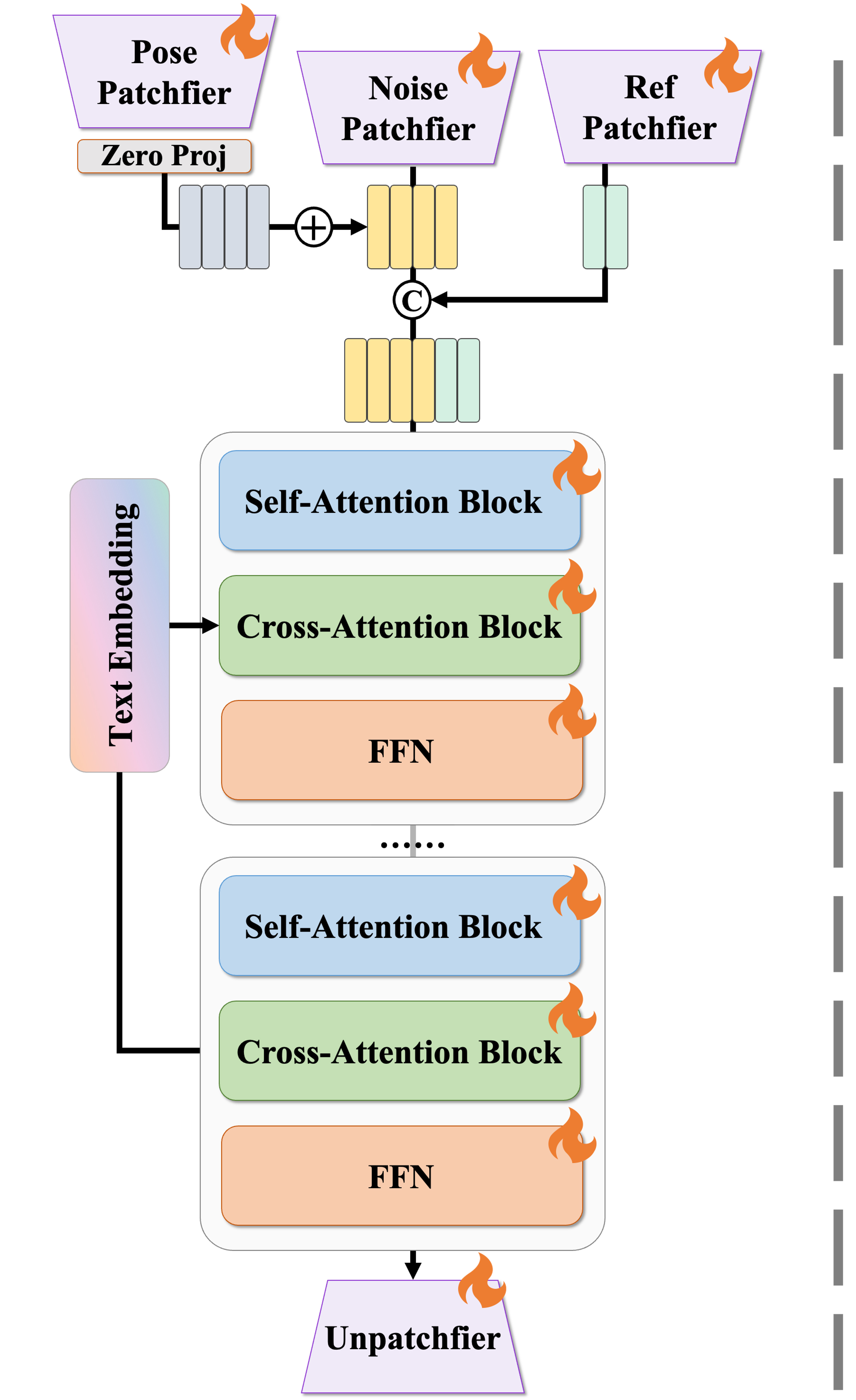} 
        \subcaption{\scriptsize{Full fine-tune}}
    \end{minipage} &
    \begin{minipage}[b]{0.22\linewidth} 
        \centering
        \includegraphics[height=65mm]{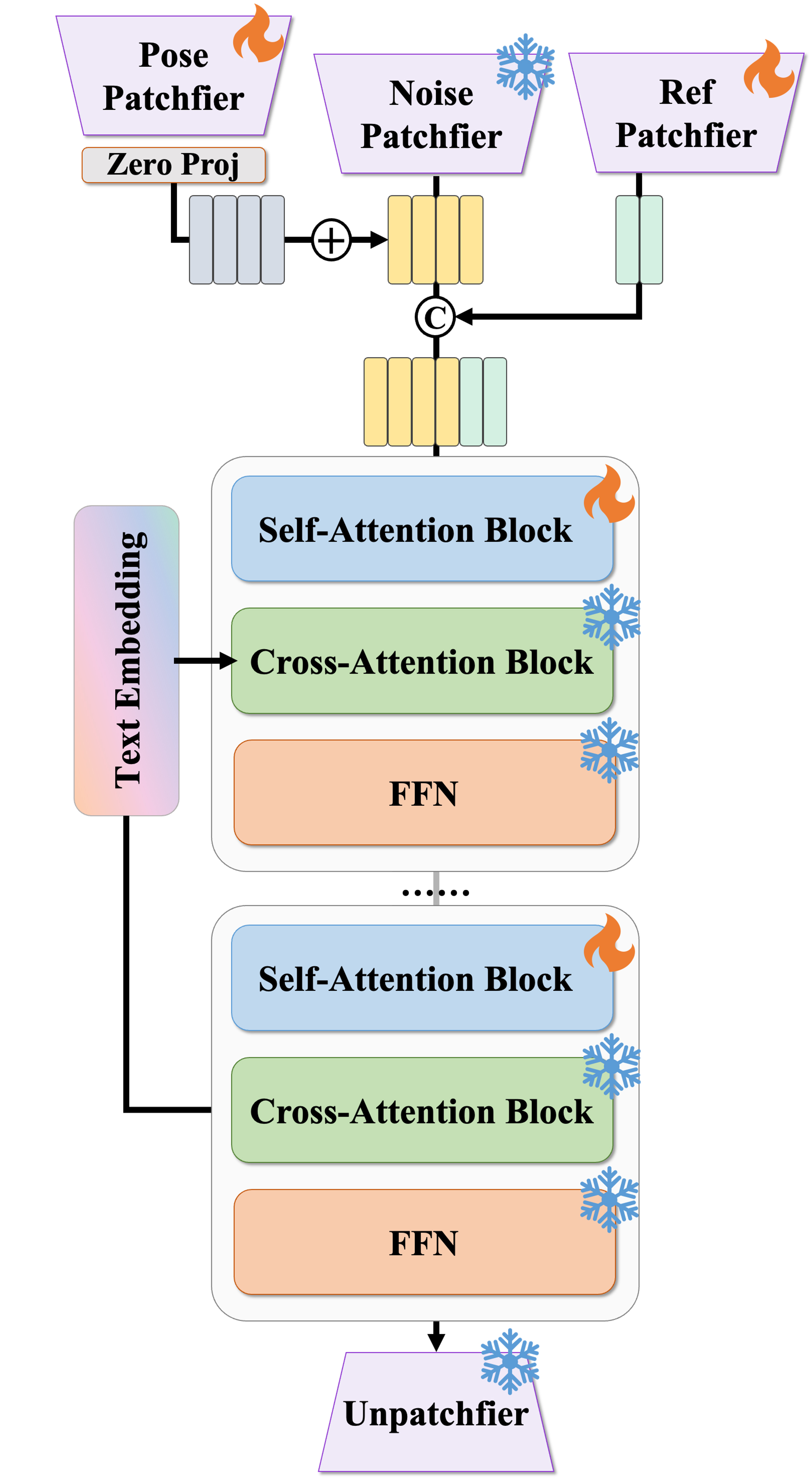} 
        \subcaption{\scriptsize{Self-attn fine-tune}}
    \end{minipage}
\end{tabular}
\caption{Illustration of architecture modifications and fine-tunable model parameters. The proposed RealisDance-DiT is fine-tuned under the final setting.}
\label{fig:model_strcture}
\end{figure*}

\section{Related work}
Controllable character animation has drawn significant attention since the GAN-based methods~\cite{siarohin2019first, siarohin2021motion, goodfellow2020generative, brock2018large}. Recently, with the advantage of Diffusion models, several methods~\cite{disco, animateanyone, champ, animatex} have made considerable progress in photorealistic generation. Specifically, DisCo~\cite{disco}, an early diffusion-based method, leverages ControlNet~\cite{controlnet} to incorporate both background and pose guidance and adopts the motion module to enhance cross-frame consistency, enabling animation generation from static reference images. However, it injects character ID through the CLIP~\cite{clip} feature, which loses lots of detailed information. As a result, DisCo struggles to maintain character consistency. Animate Anyone~\cite{animateanyone} injects character ID via the UNet-based Reference Net. While this effectively preserves identity in many cases, the method tends to fail when dealing with stylized characters, complex gestures, and large camera motions. Animate-X~\cite{animatex} introduces a pose indicator approach tailored to stylized characters. RealisDance~\cite{realisdance} integrates the 3D hand condition (HaMeR~\cite{hamer}) to improve hand fidelity.  HumanVid~\cite{humanvid} attempts to address camera motions in realistic settings.  Despite these advances, existing methods perform unsatisfactorily in open scenes, especially when encountering rare poses, complex lighting conditions, character-object interactions, and scene changes. We point out that this is attributed to the weak image-to-video main network used by existing methods. There is an urgent demand for applying powerful native video foundational models to handle challenges in open scenes. We note that several concurrent studies~\cite{omnihuman, humandit, vace} are exploring this direction. We go beyond them by proposing simple yet effective model modifications and two practical fine-tuning strategies. The proposed RealisDance-DiT is a structurally simple, empirically robust, and experimentally excellent baseline, which pushes the boundaries of controllable animation in the wild.
\section{Method}
\subsection{Simple model modifications}
We build RealisDance-DiT based on Wan-2.1. Figure~\ref{fig:model_strcture} shows several architecture modifications we explored. Initially, we attempted to directly transfer Reference Net to Wan-2.1. However, this results in an overly large network that is difficult to fine-tune given limited GPU resources. We had to remove some blocks from the Reference Net to reduce the fine-tunable parameters. However, the down-scaled Reference Net converges slowly and yields only mediocre performance. Then, we experimented with directly concatenating the reference latent to the noise latent and fine-tuning the entire network. Surprisingly, this simple design converges much quicker, and the fine-tuned model adapts to the downstream controllable character animation tasks excellently. This indicates that the large native video foundation model inherently possesses the capability to generalize to downstream tasks. The key is not to modify the model architecture but to bring out its inherent abilities. Based on this goal, we further try to only fine-tune the newly introduced condition patchifiers, the zero projection layer, and the self-attention blocks. Compared to full fine-tuning, fine-tuning with fewer parameters does not slow down convergence or degrade the final performance, which confirms our hypothesis.

There are several important details to highlight. We use the same three pose conditions as those used in RealisDance~\cite{realisdance}, \ie, HaMeR~\cite{hamer}, DWPose~\cite{dwpose}, and SMPL-CS~\cite{realisdance}. All pose conditions and the reference image are encoded using the original Wan-2.1 VAE. The encoded pose latents are concatenated along the channel dimension. Then, the concatenated pose latent and the reference latent are fed into the pose and reference patchifiers, respectively. The pose patchifier is initialized randomly, while the reference patchifier is initialized using the weights from the noise patchifier. Finally, the pose latent is added to the noise latent, and the reference latent is concatenated with the noise latent along the sequence length, before being sent to the subsequent DiT blocks.

\begin{figure}[t]
\includegraphics[width=0.95\linewidth]{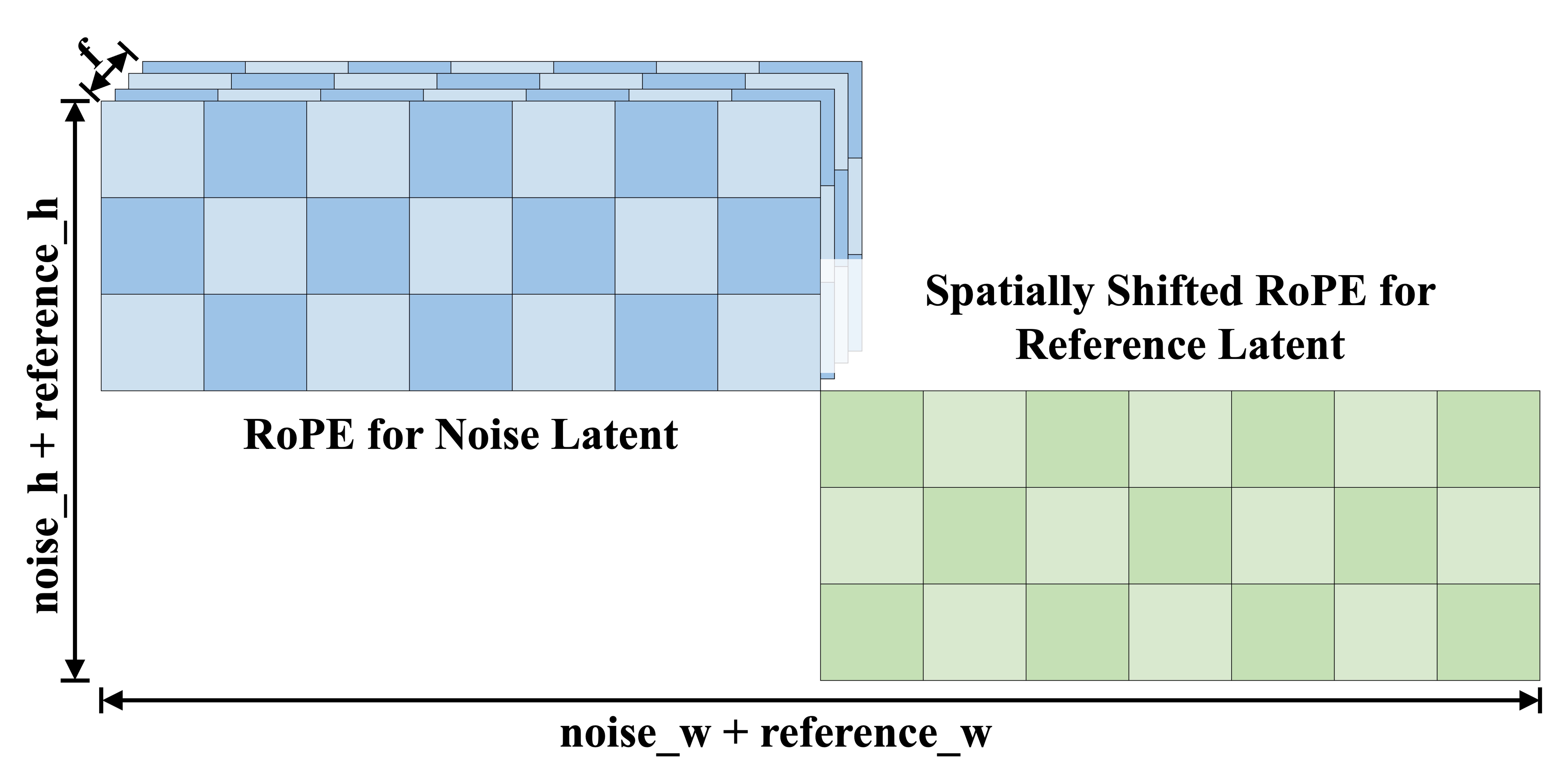}
\centering
\caption{Illustration of spatially shifted RoPE for the reference latent.}
\label{fig:shifted_rope}
\end{figure}

We also replace the Rotary Position Embedding (RoPE)~\cite{rope} used in self-attention with the shifted RoPE. As illustrated in Figure ~\ref{fig:shifted_rope}, the reference latent does not share RoPE with the noise latent. It employs the spatially shifted RoPE at the first frame, where the shifting is according to the height and width of the noise latent. 

Furthermore, our simple modifications can be seamlessly extended to the Wan-2.1 I2V model. In fact, the final version of RealisDance-DiT is developed based on Wan-2.1 I2V, where we omit the first frame, apply an all-zero mask to the noise latent, and input the reference image to the CLIP model instead of the first frame.

\subsection{Fine-tuning strategies}
We further propose the low-noise warmup and ``large batches and small iterations'' strategies for fast convergence while preserving rich priors in Wan-2.1.

\paragraph{Low-noise warmup strategy.} The timestep sampling strategy is critical for the training stability and final performance of the diffusion model. Although several methods~\cite{sd3_origin_paper, timestep_method1, timestep_method2, timestep_method3} have explored various types of sampling distributions, such as uniform distribution and logit-normal distribution, all of them utilize a fixed distribution throughout the entire training / fine-tuning period. We argue that the timestep sampling distribution should be dynamic to fit different phases of fine-tuning. For example, at the beginning of the fine-tuning process, it is ideal to sample more small timesteps to reduce the level of added noise. Samples with lower added noise are easier for the model to process, thus helping to stabilize fine-tuning in the early stages. During the middle stage of fine-tuning, the probability of sampling intermediate or larger timesteps should gradually increase, as this helps the model adapt to various timesteps in the downstream task. Based on this idea, we propose the low-noise warmup strategy, which is implemented by a dynamic sampling distribution.

\begin{figure}[t]
\includegraphics[width=0.95\linewidth]{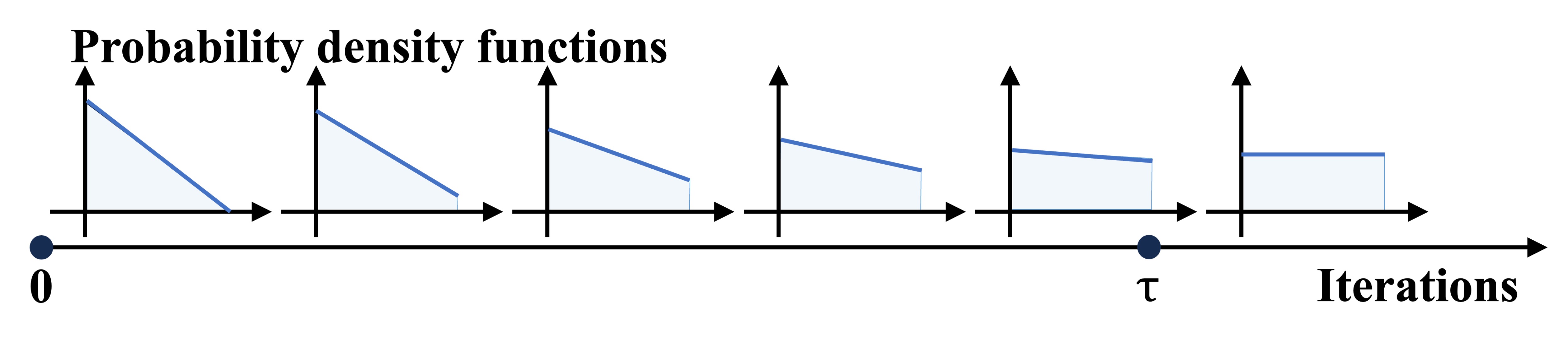}
\centering
\caption{Illustration of low-noise warmup strategy.}
\label{fig:warm_up}
\end{figure}

Given the widely used uniform sampling strategy, its probability density function can be expressed in the following formula
\begin{equation}
f(x) = 
\begin{cases} 
\frac{1}{T_\textrm{max}} & \textrm{if } 0 \leq x \leq T_\textrm{max} \\ 
0 & \textrm{otherwise} 
\end{cases},
\end{equation}
where $T_\textrm{max}$ is the maximum of the timestep, which is typically set to 1000. The proposed low-noise warmup strategy introduces an iteration-relevant component to the above probability density function, facilitating dynamic timestep sampling. Its probability density function can be expressed as below
\begin{equation}
\small
f(x) = 
\begin{cases} 
-f(i)x + \frac{T_\textrm{max}}{2}f(i) + \frac{1}{T_\textrm{max}} & \textrm{if } 0 \leq x \leq T_\textrm{max} \\ 
0 & \textrm{otherwise} 
\end{cases},
\end{equation}
where $f(i)$ can be any function related to fine-tuning iteration $i$, with a value range in $[0, \frac{2}{T_\textrm{max}^2}]$. Here we simply use a linear function, for example, which can be expressed as
\begin{equation}
f(i) = 
\begin{cases} 
\alpha\frac{2(\tau-i)}{\tau T_\textrm{max}^2} & \textrm{if } 0 \leq i \leq \tau \\ 
0 & \textrm{otherwise} 
\end{cases},
\end{equation}
where $\alpha \in [0, 1]$ is a hyperparameter that controls the maximum value of $f(i)$, $\tau$ notes the threshold of maximum warmup step. Figure~\ref{fig:warm_up} illustrates the low-noise warmup strategy. When iteration $i$ is smaller than the maximum warmup threshold $\tau$, there is a greater probability of sampling small timesteps. As the iteration $ i$ increases, the probability of sampling large timesteps rises. Once $i$ exceeds $\tau$, the sampling distribution degrades to uniform sampling.

\paragraph{Large batches and small iterations.} 
We recommend fine-tuning the model with larger batches and fewer iterations. This strategy has two benefits for fine-tuning. On the one hand, by utilizing a larger batch size, the model can be updated with more informative gradients, allowing it to focus on key factors relevant to the downstream task, without being affected by noise in the data. Therefore, the model converges faster and more stably. On the other hand, fewer iterations reduce the risk of overfitting the downstream dataset. We found that after the model adapted key factors relevant to the downstream task, it began to fit inductive biases in the downstream data. In other words, as fine-tuning continues, the loss reduction comes more from the model learning the inductive biases in the downstream data, such as high-frequency details. We also observed that, with more iterations, the controllability is slightly enhanced, but the diversity of generated results is greatly reduced, and the frequency of artifacts in the generated results increases. This means that the prior gradually disappears with the increase of fine-tuning iterations, while the prior of the foundation model is crucial for controllable character animation in open scenes.

\begin{figure*}[t]
\centering
\includegraphics[width=\linewidth]{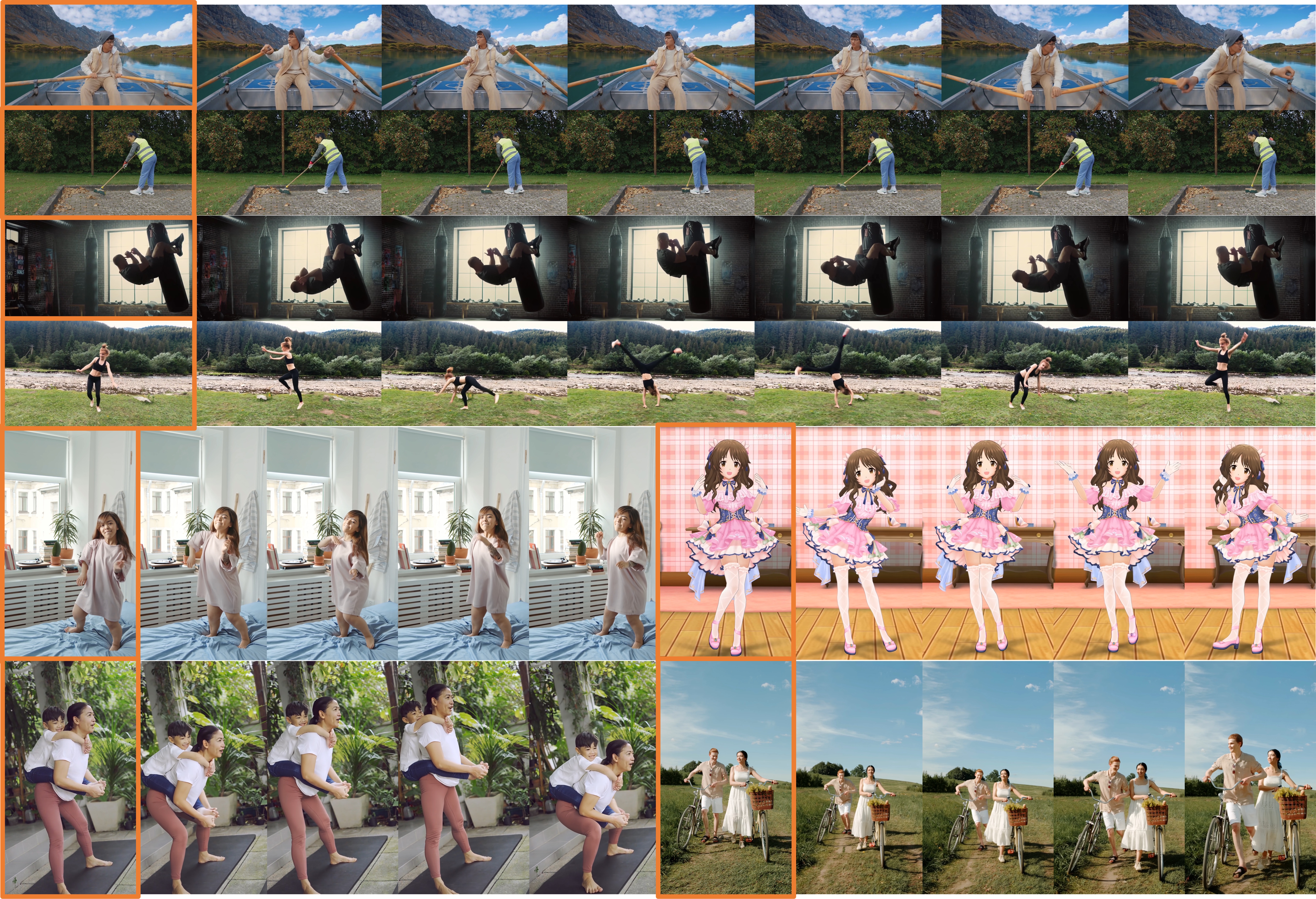}
\caption{Visualization of frames generated by RealisDance-DiT. The images with orange borders are reference images. Zoom in for better visibility. Please refer to the project page for more videos.}
\label{fig:select_results}
\end{figure*}

\subsection{Inference strategies}

During fine-tuning, we randomly drop reference images and text prompts at a rate of 5\%, allowing the model to set the CFG~\cite{cfg} scale to 2 at inference. 
To handle reference characters with diverse body shapes, we also employ an optimization-based approach to refine and replace shape parameters $\beta$ of the SMPL-X model~\cite{smplx}. Starting with an initial estimate of ${\beta}_\text{init}$ and pose obtained by GVHMR~\cite{gvhmr}, we utilize SMPLify-X to optimize $\beta$ guided by estimated 2D keypoints and human silhouettes. The objective is to minimize the discrepancy between the projected 3D silhouette and the reference silhouette while maintaining accurate alignment of the 2D keypoints. Finally, the refined shape parameters are used to replace the default shape parameters of the driving pose, enabling shape alignment between the reference characters and the pose sequence. 

\section{Experiments}

We compared RealisDance-DiT with several open source methods, including MooreAA~\cite{mooreaone}, Animate-X~\cite{animatex}, ControlNeXt~\cite{ControlNeXt}, MimicMotion~\cite{mimicmotion}, and MusePose~\cite{musepose}. The concurrent work OmniHuman~\cite{omnihuman} and HumanDiT~\cite{humandit} have not yet released their code and models, so we cannot make comparisons. The concurrent work VACE~\cite{vace} made its code public a few days before submission, which does not give us enough time to evaluate their methods.

Experiments are conducted on the TikTok dataset~\cite{tiktokdataset}, UBC fashion video dataset~\cite{ubcdataset}, and RealisDance-Val datasets, where RealisDance-Val contains 100 videos collected from the Internet, covering various characters, scenes, rare poses, lighting conditions, and character-object interactions. For TikTok dataset and UBC fashion video dataset, we follow the evaluation settings in HumanVid~\cite{humanvid}. Specifically, the method predicts frames within the range [1,72) with a stride of 3, obtains a sequence of 24 frames. The reference image is selected as the middle frame within the prediction range. The prediction resolution is set to 512$\times$896. For RealisDance-Val, the method predicts the first 5 seconds of the video at a resolution of 576$\times$1024 or 1024$\times$576 according to the original aspect ratio. We manually select the most informative frame as the reference image. 

We train two versions of RealisDance-DiT: one for ablation experiments and the other as the final version. The RealisDance-DiT used for the ablation experiments is based on the Wan-2.1 T2V 1.3B model and trained on a dataset containing 204.8k high-quality videos bought from vendors. The final version of RealisDance-DiT is based on the Wan-2.1 I2V 14B model and trained on a dataset comprising 1M high-quality videos bought from vendors. 
Both training datasets exclude the three test datasets mentioned above to ensure that no test data is used during training. During training, we employ AdamW as the optimizer and set the learning rate to 1e-5.

\begin{figure*}[t]
\centering
\includegraphics[width=0.97\linewidth]{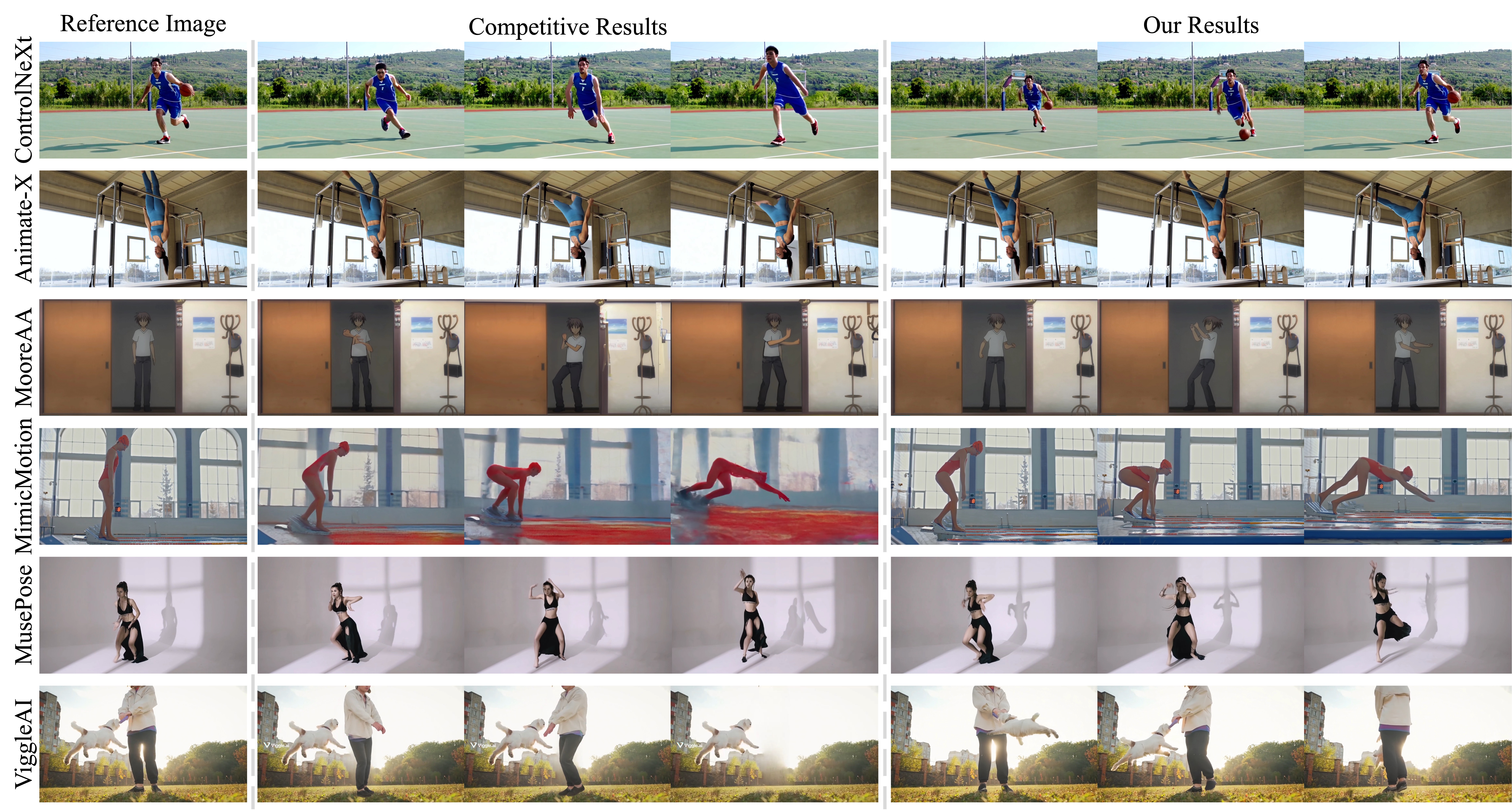}
\caption{Qualitative comparisons between RealisDance-DiT and other methods. Zoom in for better visibility. Please refer to the project page for more videos.}
\label{fig:comp_all}
\end{figure*}

\subsection{Overall comparisons}
\paragraph{Qualitative evaluation and comparisons.} Figure~\ref{fig:select_results} visualizes several video frames generated by RealisDance-DiT. As can be seen from the first two rows of the figure, RealisDance-DiT effectively handles the interactions between characters and objects. For example, in the first row, the paddle moves naturally as the character rows the boat, and in the second row, the broom moves naturally as the character sweeps the floor. The third row shows the generation capabilities of RealisDance-DiT in complex lighting scenarios, where it accurately renders light and shadow on the character according to physical principles. The fourth row evaluates the performance of RealisDance-DiT under rare poses. When the pose estimation is accurate, the model generates smooth, physically consistent videos. The fifth row demonstrates that RealisDance-DiT generalizes to characters of various body shapes and styles. Finally, the last row shows the potential of RealiDance-DiT for application in multiple character scenarios.

Figure~\ref{fig:comp_all} qualitatively compares RealisDance-DiT with existing methods and products. In the first row of Figure~\ref{fig:comp_all}, the reference image depicts a man playing basketball. ControlNeXt~\cite{ControlNeXt} effectively generates the man's movements, but the basketball disappears from the scene. In contrast, RealisDance-DiT not only successfully generates the basketball, but also ensures that the 
basketball bounces realistically after landing during dribbling. In the second row, Animate-X~\cite{animatex} fails to generate legs when facing complex yoga poses. In the third row, when dealing with animated characters, the human body parts generated by MooreAA exhibit numerous artifacts. For instance, in the third row, the model fails to accurately represent what the arms should look 
like in different poses. In contrast, the proposed method generalizes well to anime characters, thanks to the priors inherited from Wan-2.1. In the fourth row, MimicMotion fails to generate the dynamic scene caused by camera motion, while RealisDance-DiT can effectively handle camera motion and maintain detail consistency with the reference image. In the fifth row, when addressing complex lighting conditions, the shadows produced by MusePose do not align with the poses of the character. In contrast, RealisDance-DiT generates more accurate shadows that correspond to the dance movements. In the last row, we compare RealisDance-DiT with a commercial product from ViggleAI. As can be seen, the proposed method effectively handles the interaction between the character and the dog when the character spins. However, in the results generated by ViggleAI, the dog is always suspended in the air, and the character disappears when she turns back.

\begin{table*}[h]
\centering
\small
\tabcolsep=1mm
\begin{tabular}{lcccccccccc}
    \toprule
    \multirow{2}{*}{Method} & {I2V} & {I2V} & {Subject} & {BG} & {Temporal} & {Motion} & {Dynamic} & {Aesthetic} & \multirow{2}{*}{FVD$\downarrow$} & \multirow{2}{*}{FID$\downarrow$} \\
    & {Subject}$\uparrow$ & {BG}$\uparrow$ & {Consist}$\uparrow$ &  {Consist}$\uparrow$ & {Flicker}$\uparrow$ & {Smooth}$\uparrow$ & {Degree}$\uparrow$ & {Quality}$\uparrow$ & & \\
    \midrule
    Animate-X   & \textbf{96.06} & \textbf{96.59} & 93.83 & 94.83 & 97.40 & 98.52 & 53 & 55.22 & 855.87 & \underline{34.32} \\
    ControlNeXt & 92.91 & 93.92 & 91.41 & 93.57 & 96.91 & 98.05 & 63 & 55.57 & 810.82 & 37.24 \\
    MimicMotion & 92.79 & 93.80 & 91.10 & 93.20 & 96.78 & 98.20 & 59 & 53.31 & \underline{783.55} & 40.19 \\
    MooreAA     & 92.33 & 93.35 & 93.12 & 93.77 & 95.20 & 96.74 & \textbf{68} & 56.08 & 867.48 & 35.50 \\
    MusePose    & 92.24 & 93.01 & \underline{93.88} & \underline{94.88} & \textbf{97.88} & \underline{98.57} & 57 & \underline{56.28} & 1049.06 & 42.02 \\
    \midrule
    RealisDance-DiT & \underline{95.97} & \underline{96.57} & \textbf{93.91} & \textbf{95.83} & \underline{97.76} & \textbf{98.71} & \underline{66} & \textbf{57.93} & \textbf{563.28} & \textbf{24.79} \\
    \bottomrule
\end{tabular}
\caption{Quantitative Results on the RealisDance-Val. RealisDance-DiT ranks either first or second across all evaluation metrics. Especially for FVD and FID, RealisDance-DiT outperforms all compared methods by a large margin.}
\label{tbl:realisdance_val}
\end{table*}

\begin{table}[t]
\centering
\tabcolsep=0.5mm
\small
\begin{tabular}{lccccc}
    \toprule
    Method & SSIM$\uparrow$ & PSNR$\uparrow$ & LPIPS$\downarrow$ & FVD$\downarrow$ & FID$\downarrow$ \\
    \midrule
    Animate-X   & 0.7427 & 16.71 &  0.2854 & 508.63 & \underline{32.77}\\
    ControlNeXt & 0.7282 & 16.31 &  0.2958 & 548.01 & 33.48\\
    MimicMotion & 0.7507 & \textbf{19.30} & \textbf{0.2203} & \underline{472.51} & 34.88\\
    MooreAA   & \textbf{0.7636} & \underline{18.62} &  \underline{0.2296} & 501.22 & 37.28\\
    MusePose & \underline{0.7566} & 18.20 & 0.2484 & 532.75 & 41.99\\
    \midrule
    RealisDance-DiT & 0.7170  & 17.55 & 0.2613 & \textbf{458.81} & \textbf{30.39}\\
    \bottomrule
\end{tabular}
\caption{Quantitative Results on the TikTok dataset.}
\label{tbl:tt}
\end{table}

\begin{table}[t]
\centering
\small
\tabcolsep=0.5mm
\begin{tabular}{lccccc}
    \toprule
    Method & SSIM$\uparrow$ & PSNR$\uparrow$ & LPIPS$\downarrow$ & FVD$\downarrow$ & FID$\downarrow$ \\
    \midrule
    Animate-X   & 0.8931 & 22.15 &  0.0691 & \textbf{70.47} & \textbf{10.11}\\
    ControlNeXt & 0.8530 & 18.48 &  0.1320 & 143.02 & 13.82\\
    MimicMotion & \textbf{0.9126} & \textbf{23.80} & \underline{0.0605} & 80.89 & 15.40\\
    MooreAA   & 0.8795 & 20.83 &  0.0929 & 149.66 & 21.74\\
    MusePose & 0.8955 & 22.20 &  0.0665 & 96.17 & 14.95\\
    \midrule
    RealisDance-DiT & \underline{0.9083} & \underline{23.33} & \textbf{0.0526} & \underline{72.94} & \underline{10.81}\\
    \bottomrule
\end{tabular}
\caption{Quantitative Results on the UBC fashion video dataset.}
\label{tbl:ubc}
\end{table}

\paragraph{Quantitative comparisons.} For all quantitative evaluations, we select the same frames across different methods to calculate FID and FVD, ensuring fairness in the comparisons. 
Table~\ref{tbl:tt} shows the quantitative evaluation on the TikTok dataset. It is worth noting that, as we follow the setting in HumanVid~\cite{humandit}, we use the last 40 videos from the TikTok dataset out of a total of 340 videos as the test set. Therefore, for several methods like MooreAA, these data could be included in their training data. Even under such an unfair comparison, RealisDance-DiT achieves the best FVD and FID values among the compared methods. 
Table~\ref{tbl:ubc} presents the comparison results on the UBC fashion video dataset. Since the UBC fashion video dataset is not utilized as training data by any other methods, this comparison is more equitable. RealisDance-DiT achieves competitive results, ranking either first or second across all evaluation metrics.

The above two datasets are too na\"{\i}ve to effectively evaluate the model capabilities in real-world scenarios. Therefore, we have collected a new test dataset, RealisDance-Val, to evaluate existing methods. Unlike the previous settings, we do not use low-level metrics such as SSIM, PSNR, and LPIPS, because our test set encompasses a wider range of open scenes, where the background of some test videos may change. Consequently, the model must generate new content based on the background of the reference images. However, since the background occupies the majority of the frame, the difference between the newly generated background and the ground truth will result in low values for those low-level metrics, even if the new content is very realistic. Therefore, we utilize Vbench-I2V~\cite{vbench_i2v} as the metric for the RealisDance-Val dataset. Table~\ref{tbl:realisdance_val} exhibits the comparison results on the RealisDance-Val dataset. RealisDance-DiT ranks either first or second across all evaluation metrics. Especially for FVD and FID, RealisDance-DiT outperforms all compared methods by a large margin.



\begin{table}[t]
\centering
\small
\tabcolsep=1.2mm
\begin{tabular}{lcccc}
    \toprule
     & Ref. Net & Light Ref. Net & Full Ft. & Part Ft.\\
    \midrule
    FID & OOM & 31.01 & \underline{25.58} & \textbf{24.79}     \\
    FVD & OOM & 678.98 & \textbf{519.22}  & \underline{563.28} \\
    \bottomrule
\end{tabular}
\caption{Comparisons between different fine-tuning designs.}
\label{tbl:model_ablation}
\end{table}

\subsection{Ablation studies}
\paragraph{Model designs.} We conducted ablation experiments to investigate the architectural designs and fine-tunable parameters of RealisDance-DiT, which is built on the Wan-2.1 I2V 14B model. Table~\ref{tbl:model_ablation} compares four settings: a Reference Net variant, a lightweight Reference Net variant, simple modifications with full fine-tuning, and simple modifications with part fine-tuning. The Reference Net variant is too heavy to be fine-tuned using limited GPU resources. Therefore, we pruned some blocks from the Reference Net. Specifically, each block in the Reference Net corresponds to every five blocks in the main network. We observed that such a lightweight Reference Net variant converges slowly, even with extra fine-tunable parameters, which achieves only 31.01 in FID and 678.98 in FVD. While simple modifications, whether through full fine-tuning or part fine-tuning, can lead to stronger baselines. 
This is because the current large video foundation models are already powerful enough, which can easily adapt to downstream tasks using their built-in priors. We should make as few modifications as possible and bring out its capabilities for downstream tasks. 
Furthermore, as demonstrated, part fine-tuning will not degrade the final performance. This is because the foundation model is large enough that fine-tuning a subset of the parameters is adequate to adapt to downstream tasks.

\begin{figure}[t]
\includegraphics[width=0.9\linewidth]{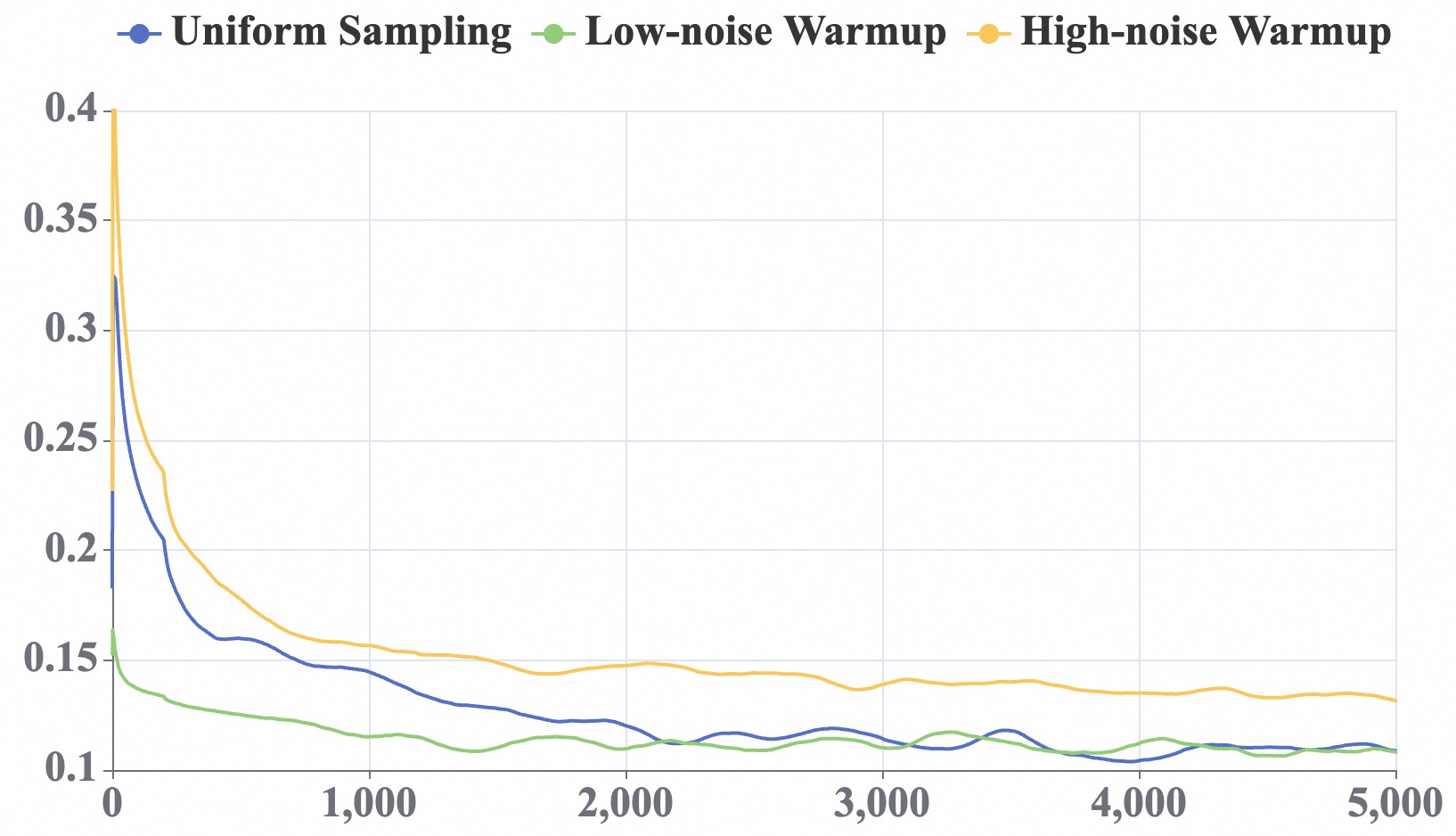}
\centering
\caption{Visualization of smoothed training loss curves.}
\label{fig:low_noise}
\end{figure}

\paragraph{Low-noise warmup strategy.} We compare the low-noise warmup strategy against fixed uniform timestep sampling and a high-noise warmup counterpart. For this comparison, we train RealisDance-DiT based on the Wan-2.1 T2V 1.3B model. Figure~\ref{fig:low_noise} illustrates the smoothed training loss curves. As shown, the low-noise warmup strategy accelerates convergence compared to the standard fixed uniform timestep sampling. Additionally, Figure~\ref{fig:low_noise} demonstrates that the high-noise warmup counterpart slows down convergence, further confirming our hypothesis that low-noise samples are crucial for stabilizing the early stages of fine-tuning.

\begin{figure}[t]
\includegraphics[width=0.95\linewidth]{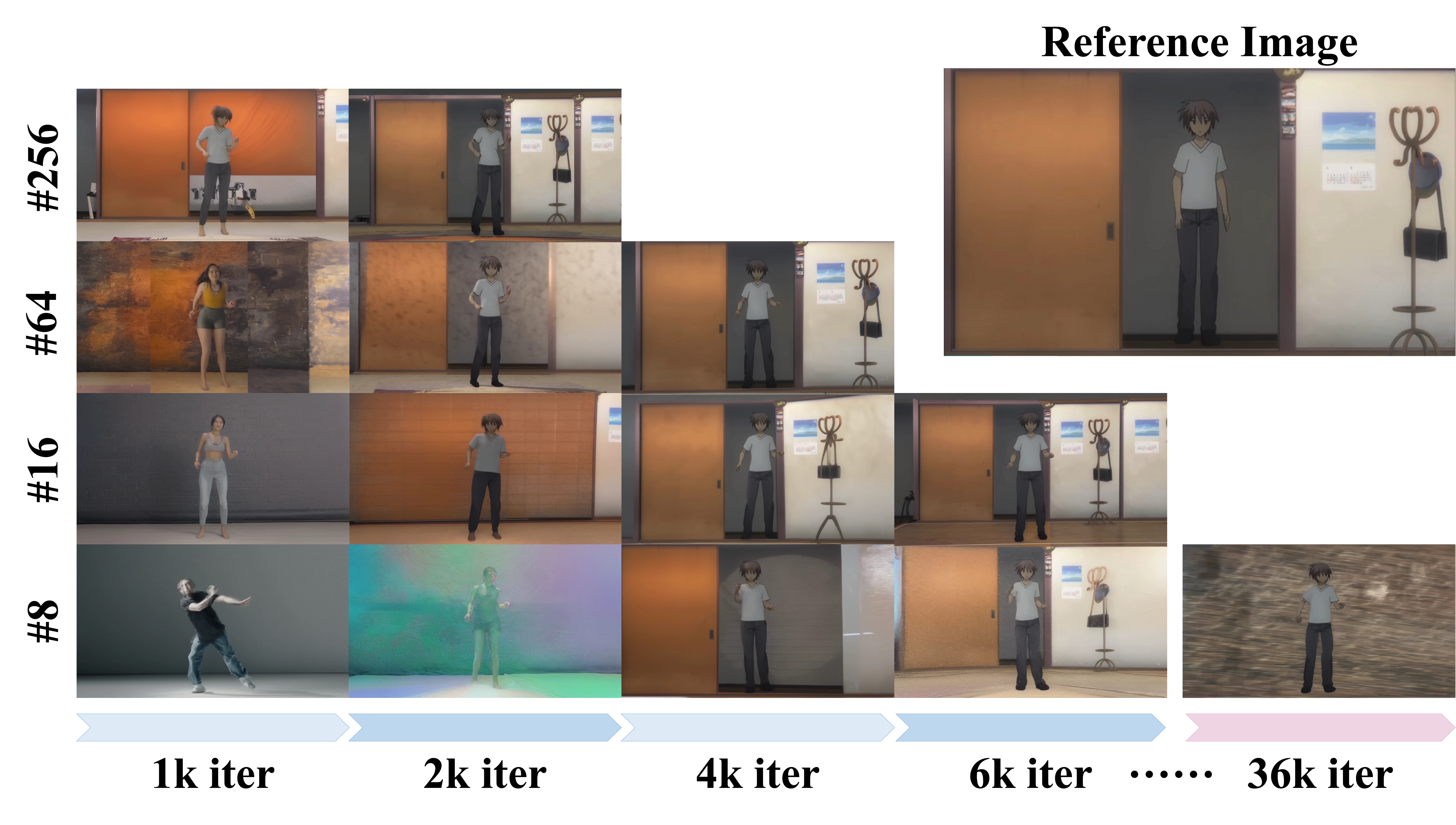}
\centering
\caption{Visualization of different batch configurations on the RealisDance-Val dataset. `\#' denotes the batch size. Zoom in for better visibility.}
\label{fig:bvi}
\end{figure}

\paragraph{Large batches and small iterations.} We evaluate various batch and iteration settings for RealisDance-DiT, utilizing the Wan-2.1 T2V 1.3B model as the foundation for this experiment. This evaluation aims to determine how different batch configurations impact convergence speed. Due to limited GPU resources, we implement large batches using gradient accumulation. Figure~\ref{fig:bvi} visualizes the generated frames corresponding to different batch configurations at various iterations. We can see that larger batches get faster convergence speed, even with the same learning rate. Furthermore, we observed that as iterations increase, the conditional controllability slightly improves, however, the diversity of the generated results significantly decreases, and numerous artifacts are introduced. As shown in the figure~\ref{fig:bvi}, the model fine-tuned with a batch size of 8 loses the ability to preserve the background at iteration 36k, causing the background to completely transform into artifacts. This is because fine-tuning with too many iterations will make the model overfit the downstream data and lose its original prior. Therefore, we suggest using a large batch size along with a small number of iterations for fine-tuning, which facilitates rapid convergence while maximally preserving the prior knowledge of the foundation model.

\section{Limitations}
\begin{figure}[t]
\includegraphics[width=0.95\linewidth]{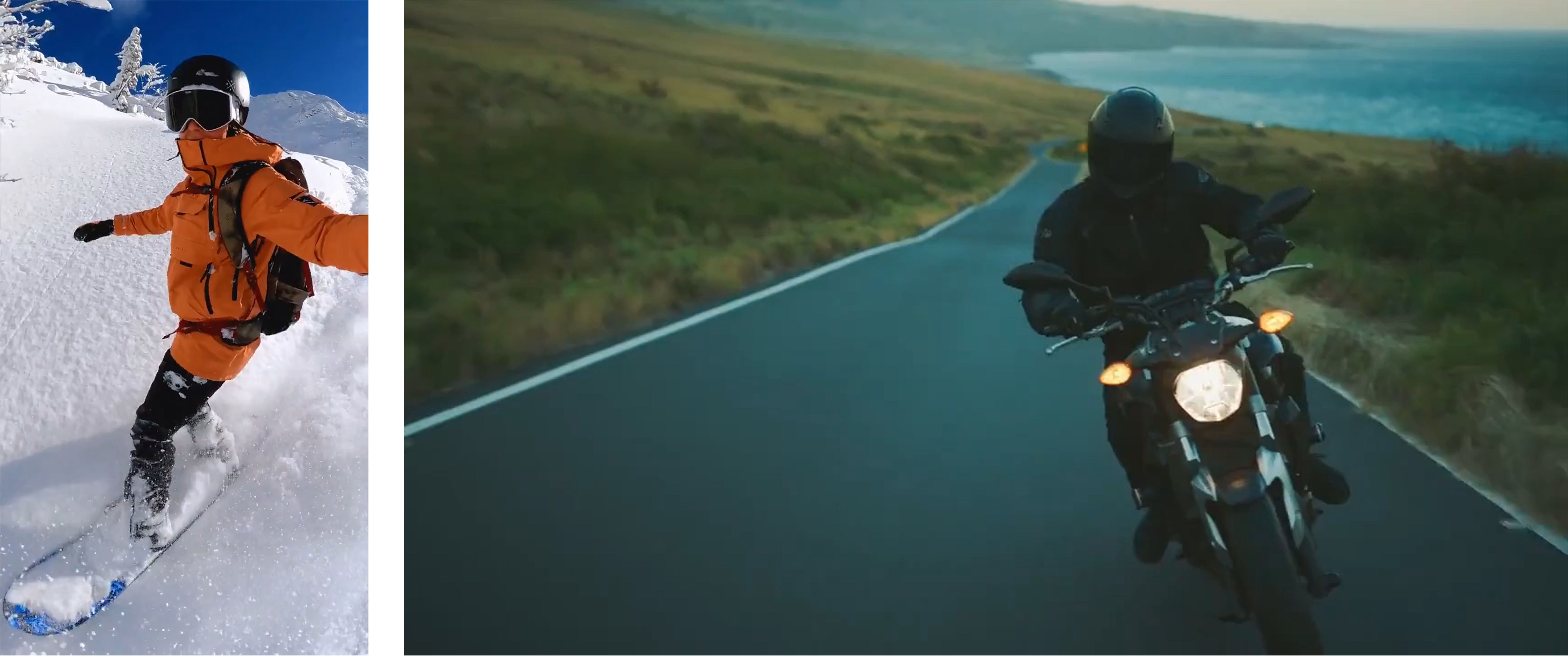}
\centering
\caption{Illustration of limitation cases. RealisDance-DiT tends to generate a static background when the character and the camera are relatively stationary.}
\label{fig:limit}
\end{figure}

There are two cases where RealisDance-DiT may not produce satisfactory results. The first case is when all three estimated poses are incorrect for extremely complex poses. In this case, RealisDance-DiT tends to generate random poses and introduces artifacts. The second case is when the character and the camera are relatively stationary, for example, the character is skiing with a GoPro or riding a motorcycle towards the camera, see the illustration in Figure~\ref{fig:limit}. In this case, RealisDance-DiT tends to generate a static background instead of a background that gradually recedes. These issues need to be addressed in future work.

\section{Conclusion}
In this paper, we present RealisDance-DiT, a simple yet powerful baseline that makes progress towards controllable character animation in the wild. 
We emphasize that as long as the foundation model is powerful enough, straightforward model modifications with flexible fine-tuning strategies can yield a superior baseline. This is because such a large native video foundation model inherently can generalize to downstream tasks. Furthermore,
we propose two fine-tuning strategies that speed up the model convergence while maximally preserving the built-in priors. Experiments are conducted on the TikTok dataset, the UBC fashion video dataset, and the proposed RealisDance-Val datasets. Both qualitative and quantitative experimental results demonstrate that RealisDance-Val significantly outperforms the other compared methods, qualifying it as a solid baseline for future research.

\bibliographystyle{plain}
\bibliography{reference}


\end{document}